\documentclass[10pt]{article} 

\usepackage[preprint]{rlj}           

\usepackage{amssymb}            
\usepackage{mathtools}          
\usepackage{mathrsfs}           
\usepackage{graphicx}           
\usepackage{subcaption}         
\usepackage[space]{grffile}     
\usepackage{url}                
\usepackage{lipsum}             

\usepackage{xspace}
\usepackage{hyperref}
\usepackage{booktabs}
\usepackage{algpseudocode}
\usepackage{fdsymbol}
\usepackage{multirow}
\usepackage{subcaption}
\usepackage{colortbl}
\usepackage{listings}
\usepackage{fancybox}
\usepackage{graphicx}
\usepackage[nameinlink]{cleveref}
\usepackage[title]{appendix}

\newcommand{\ludii}{\texttt{Ludii}\xspace}
\newcommand{\pgx}{\texttt{PGX}\xspace}
\newcommand{\methodname}{\texttt{Ludax}\xspace}

\newcommand{\syntax}[1]{{\texttt{\textcolor{violet}{#1}}}}

\lstdefinelanguage{LUDII}
{
  sensitive=false,    
  morecomment=[l]{;}, 
  alsoletter={:,-,?},   
  morekeywords={game,players,equipment,rules,start,play,end,if,is,then,no,if:,to,or,from, repeat},
}

\definecolor{SyntaxBlue}{HTML}{569CD6}
\definecolor{VariableBlue}{HTML}{9CDCFE}
\definecolor{CommentGreen}{HTML}{008000}
\definecolor{CommentGray}{HTML}{969696}
\definecolor{VariableGreen}{HTML}{009900}
\title{Ludax: A GPU-Accelerated Description
Language for Board Games}

\setrunningtitle{Ludax: A GPU-Accelerated Description Language for Board Games}

\author{%
  Graham Todd$^{\diamondsuit}$,
  Alexander G. Padula$^{\spadesuit}$,
  Dennis J.N.J. Soemers$^{\clubsuit}$,
  Sam Earle$^{\diamondsuit}$,
  Julian Togelius$^{\diamondsuit}$
}

\emails{gdrtodd@nyu.edu}

\affiliations{
$^{\diamondsuit}$\textbf{New York University Tandon}\\
$^{\spadesuit}$\textbf{ETH Zurich}\\
$^{\clubsuit}$\textbf{Maastricht University}
}

\contribution{
    We introduce \methodname, a new game description language for board games which automatically compiles into GPU-accelerated environments suitable for reinforcement learning
    }
    {
    Inspired by previous separate efforts in both game description languages (e.g. \ludii \citep{Piette_2020_Ludii}) and hardware acceleration for board games (e.g. \pgx \citep{koyamada2023pgx})
    }

\contribution{
    We demonstrate that \methodname achieves competitive throughput and training speed with hand-crafted JAX environments while supporting a wider space of games than existing libraries
    }
    {
    We compare speeds to \pgx for games which are supported by both systems and measure throughput on additional game exemplars not covered by any existing JAX implementations
    }

\contribution{
    We describe a general procedure for compiling description languages into JAX-compliant programs which could be adapted to other domains and problem settings
    }
    {
    \methodname provides an alternative approach for compiling into JAX programs compared to extant description languages that exist for other domains \citep{chandra2025memo, gimelfarb2024jaxplan, earle2025puzzlejaxbenchmarkreasoninglearning}
    }

\keywords{Games, hardware acceleration, description languages} 

\summary{Games have long been used as benchmarks and testing environments for research in artificial intelligence. A key step in supporting this research was the development of \textit{game description languages}: frameworks that compile domain-specific code into playable and simulatable game environments, allowing researchers to generalize their algorithms and approaches across multiple games without having to manually implement each one. More recently, progress in reinforcement learning (RL) has been greatly aided by advances in \textit{hardware acceleration}. Libraries like JAX allow practitioners to take full advantage of cutting-edge computing hardware, often speeding up training and testing by orders of magnitude. Here, we present a synthesis of these strands of research: a domain-specific language for board games which automatically compiles into hardware-accelerated code. Our framework, \methodname, combines the generality of game description languages with the speed of modern parallel processing hardware and is designed to fit neatly into existing deep learning pipelines. We envision \methodname as a tool to help accelerate games research generally, from RL to cognitive science, by enabling rapid simulation and providing a flexible representation scheme. We present a detailed breakdown of \methodname's description language and technical notes on the compilation process, along with speed benchmarking and a demonstration of training RL agents.
}

\begin{document}

\maketitle  

\begin{abstract}
Games have long been used as benchmarks and testing environments for research in artificial intelligence. A key step in supporting this research was the development of \textit{game description languages}: frameworks that compile domain-specific code into playable and simulatable game environments, allowing researchers to generalize their algorithms and approaches across multiple games without having to manually implement each one. More recently, progress in reinforcement learning (RL) has been greatly aided by advances in \textit{hardware acceleration}. Libraries like JAX allow practitioners to take full advantage of cutting-edge computing hardware, often speeding up training and testing by orders of magnitude. Here, we present a synthesis of these strands of research: a domain-specific language for board games which automatically compiles into hardware-accelerated code. Our framework, \methodname, combines the generality of game description languages with the speed of modern parallel processing hardware and is designed to fit neatly into existing deep learning pipelines. We envision \methodname as a tool to help accelerate games research generally, from RL to cognitive science, by enabling rapid simulation and providing a flexible representation scheme. We present a detailed breakdown of \methodname's description language and technical notes on the compilation process, along with speed benchmarking and a demonstration of training RL agents.
\end{abstract}

\section{Introduction}

For the past 75 years, games have served as vital tests and benchmarks for artificial intelligence research. While many specific games have been completely solved \citep{schaeffer2007checkers} or optimized beyond the abilities of the strongest human players \citep{campbell2002deep, silver2017mastering}, the general space of games remains a fertile ground for measuring improvements in reasoning, planning, and strategic thinking. A critical part of this progress, however, is the ability to test approaches and algorithms on a set of environments that are both diverse and computationally efficient.

To help drive further games and learning research, we introduce \methodname: a domain-specific language for board games that compiles into GPU-accelerated code written in the JAX library \citep{jax2018github}. \methodname draws on two main inspirations: (1) \ludii \citep{Piette_2020_Ludii}, a general purpose description language for board games already used to represent more than 1400 games from throughout history and around the world, and (2) \pgx \citep{koyamada2023pgx}, a collection of optimized JAX-native implementations of classic board games and video games designed to facilitate rapid training and evaluation of modern reinforcement learning (RL) agents.
\methodname presents a flexible and general-purpose game representation format that can be leveraged for efficient simulation and learning on modern computing hardware.

\methodname currently supports two-player, perfect-information, turn-based board games played by placing, capturing, and moving pieces. This set of mechanics is broad enough to capture a wide range of existing games (e.g. \textit{Connect Four}, \textit{Pente}, \textit{Hex}, ...) as well as many
unexplored \textit{novel} games and variants that fall within that class.
Further, \methodname is designed to be easily expandable -- like with \ludii, implementing new game mechanics in \methodname only requires implementing new atomic components in the underlying description language. These components can then be combined compositionally with existing elements of the language to produce an entirely new \textit{range} of possible games, instead of each game needing to be implemented separately. 

\begin{figure}[t]
    \centering
    \includegraphics[width=\linewidth]{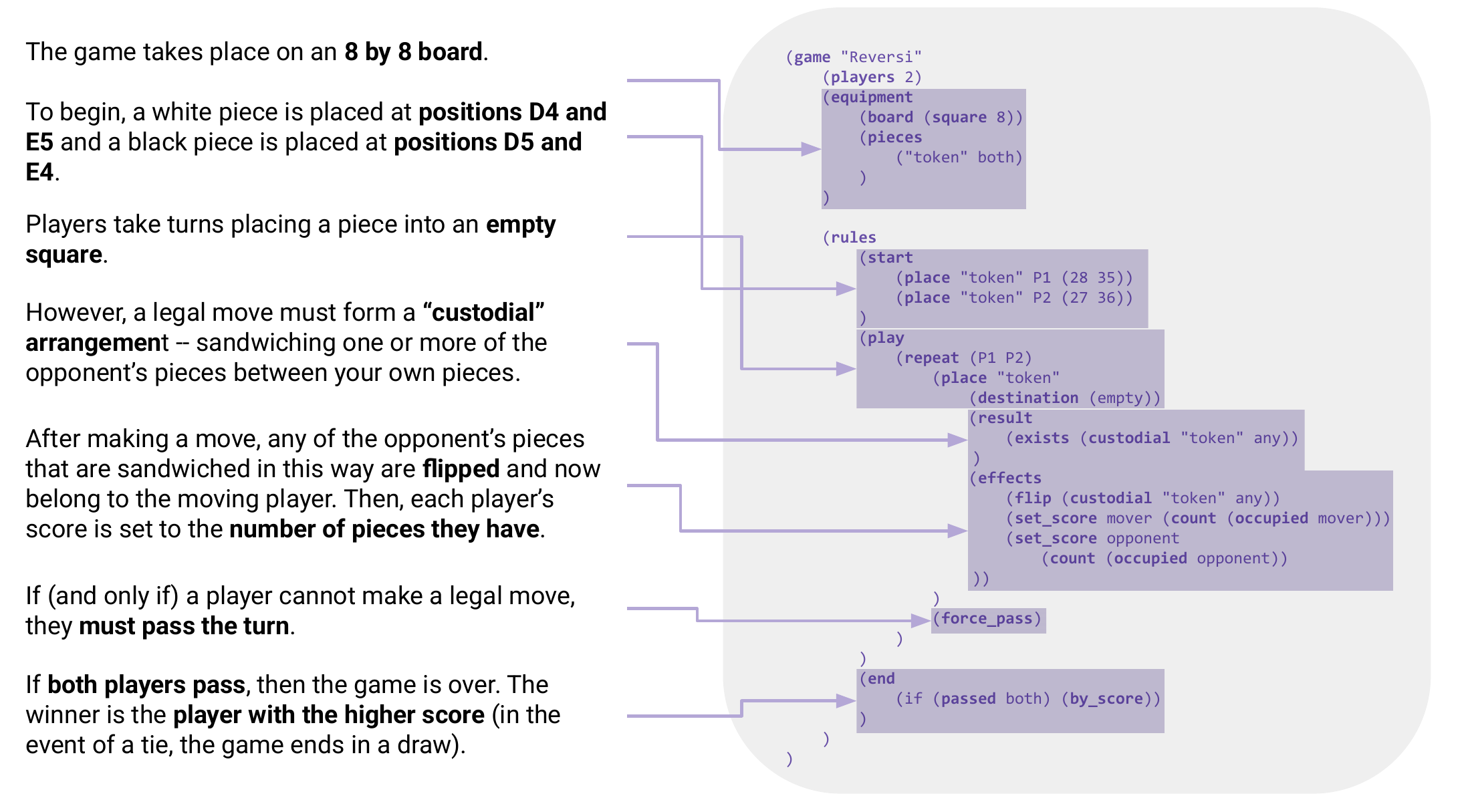}
    \caption{\textbf{Natural language description of \textit{Reversi} along with its corresponding translation into \methodname}. \methodname uses ``ludemic'' syntax that represents high-level game components as separate program sections and aims to be easily interpretable to non-experts.}  
    \label{fig:dsl-example}
\end{figure}

Another design goal for \methodname is ease of use, both in terms of game design and experimentation. The syntax of the description language is ``ludemic'' \citep{Piette_2020_Ludii} -- splitting game rules into clear sections governing the game's setup, play mechanics, and end conditions. Like with \ludii, game programs in \methodname resemble English descriptions of rules (see \autoref{fig:dsl-example}). Further, by leveraging the structure of the existing \pgx library, environments instantiated in \methodname can be easily combined with existing frameworks for GPU-accelerated search, reinforcement learning, or evolution \citep{deepmind2020jax, tang2022evojax}. \methodname also supports a basic web interface for interactive debugging and potential user-studies.

\methodname is fundamentally a platform for accelerating board game research. In an era of increasingly complicated tasks and benchmarks, relatively simple board games may seem to be less interesting research domains (especially as many games have been more-or-less ``solved'' by modern methods). However, \methodname is not just a collection of new tasks. By decoupling rapid execution from the intensive process of writing new environment code, \methodname can power new research in a variety of directions. For instance, \methodname can be used to analyze RL generalization \citep{soemers2025environment} by defining a wide range of modifications for a target game task (akin to a platform like \texttt{Minihack} \citep{Samvelyan_2021_MiniHack}) or help improve studies of game generation by enabling the rapid evaluation of procedurally-generated rulesets \citep{todd2024gavel, collins2025generation}. Finally, \methodname can help advance recent research into world modeling \citep{ying2025assessing} by (1) providing a wide and easily-refreshable set of environments to test on efficiently and (2) allowing automated systems to propose and refine world models in these ``novel games'' by writing high-level and semantically-meaningful DSL code. 

To our knowledge, \methodname is the first board game description language which compiles into GPU-accelerated code. In the following sections, we provide a detailed description of the language syntax, compilation process, and \methodname's expressive range. We also provide speed benchmarking compared to both \ludii and \pgx, as well as an initial demonstration of training learned agents. Finally, we conclude with a discussion of potential use cases and future directions. \methodname is open-source under the Apache 2.0 license, and the code is available at: \url{https://github.com/gdrtodd/ludax}.


\section{Related Work}


\textbf{Game Description Languages: } Game description languages have been used for many years and in a variety of domains. The Stanford GDL \citep{Love_2008_GDL,Genesereth_2014_GGP,Schiffel_2014_Representing,Thielscher_2017_GDLIII} is among the most influential, helping to popularize research in general game playing \citep{Pitrat_1968_GGP} through its use in the International General Game Playing Competition \citep{Genesereth_2005_GGP, Genesereth_2013_GGP}. Other notable examples include \texttt{VGDL} \citep{ebner2013towards,Schaul_2013_VGDL,Schaul_2014_Extensible} (primarily known from its use in the General Video Game AI framework \citep{Perez_2019_GVGAI}), \texttt{RBG} \citep{Kowalski_2019_Regular}, \texttt{Ludi} \citep{Browne_2009_PhD}, and its successor \ludii \citep{Piette_2020_Ludii}. GDLs have also been used to describe the rules of card games \citep{font2013card} as well as to represent human goals in naturalistic simulated environments \citep{davidson2022creativity, davidson2025goals}. Modern game description languages have tended to move away from a basis in formal logic in favor of greater human usability, though there are benefits in efficiency gained by the use of regular languages \citep{Kowalski_2020_Efficient}.


\textbf{GPU-Accelerated Environments: } Recent years have seen a proliferation of learning environments implemented in the JAX library or other frameworks that enable hardware (typically GPU) acceleration. Examples include single-agent and multi-agent physics simulators \citep{freeman1brax, Makoviychuk_2021_Isaac, Bettini_2022_VMAS}, ports of both classic and recent reinforcement learning tasks \citep{Dalton_2020_Accelerating, gymnax2022github, koyamada2023pgx, matthews2024craftax}, combinatorial optimization problems \citep{bonnet2023jumanji}, multi-agent coordination problems \citep{rutherford2023jaxmarl}, and driving simulators \citep{Gulino_2023_Waymax, kazemkhani2025gpudrive}. While these efforts have spurred significant progress and span a wide range of domains and task formulations, each of them implement a fixed environment or set of environments. As such, they cannot easily be extended to novel environments without first writing new hardware-accelerated code. \methodname stands alongside a number of description languages for other domains (e.g. probabilistic programming, planning, single-player puzzles) that leverage JAX for efficient execution \citep{chandra2025memo, gimelfarb2024jaxplan, earle2025autoverse, earle2025puzzlejaxbenchmarkreasoninglearning}.

\section{Description Language Details}

\methodname's game description language draws heavily on the \ludii description language, particularly in its use of ``ludemic'' syntax that represents game rules in terms of high-level and easily-understandable components \citep{Piette_2020_Ludii}. The complete grammar file and syntax details are available in the Supplemental Material.

\subsection{Equipment and Start Rules}

The \textcolor{violet}{\texttt{equipment}} section contains information about the physical components used by the game. This includes the size and shape of the board (e.g. whether it is rectangular or hexagonal), the piece types used by players during the game, and optionally any ``board regions'' that are referred to by later rules (e.g. particular tiles that are blocked off from play).
The \textcolor{violet}{\texttt{start}} section is an optional section that contains the rules for the game's setup. For most games, play begins on an empty board and the \textcolor{violet}{\texttt{start}} section is omitted. In some games, such as \textit{Reversi} (see \autoref{fig:dsl-example}), pieces are placed in a particular arrangement at the start of play.

\subsection{Play Rules}

Typically, the \textcolor{violet}{\texttt{play}} rules of each game are the most involved, as they detail the core mechanics and dynamics of the game. The \textcolor{violet}{\texttt{play}} section is itself broken into one or more subsections called ``play phases.'' Each phase has its own rules for player actions and turn-taking, as well as specific conditions for when to transition to another phase. Most games have only a single phase in which players alternate turns until the game is over, though \methodname also supports games with more complicated dynamics.

The core of each phase is a ``play mechanic'' that encodes the ways that players take their turns. In the context of reinforcement learning, a play mechanic specifies both the action space ($\mathcal{A}$) and the transition function ($\mathcal{T}: \mathcal{S} \times \mathcal{A} \rightarrow \mathcal{S})$. At a lower level, each play mechanic also defines a ``legal action mask function'' that returns whether each action is valid from the current game state. \methodname supports two kinds of play mechanics: \syntax{place} and \syntax{move}, the latter of which is broken down into \syntax{step}, \syntax{hop}, and \syntax{slide}. Each play mechanic takes arguments that define exactly how it is implemented. For instance, a \syntax{place} rule takes a \syntax{destination} constraint which can be used to restrict where pieces can be placed, as in \textit{Connect Four}. Arguments to \syntax{hop} can be used to specify which pieces can be hopped over and in which directions, and the different movement types can be composed to create intricate rulesets like \textit{English Draughts} (see \autoref{sec:dsl-examples}). Finally, a play mechanic might optionally define one or more \syntax{effects} that modify the game state after the action is performed.
Effects are used to handle mechanics like capturing or flipping pieces, as well as updating each player's score if the game uses it -- see \textit{Reversi} again for an example.

High-level operators like \syntax{place} and \syntax{slide} eventually ground out to low-level components that encode individual game mechanics. These components, referred to as \syntax{masks}, \syntax{functions}, and \syntax{predicates} based on their return values, take in the current game state and compute a single, specific game property. For instance, the commonly-used \syntax{line} function returns the number of contiguous lines of a particular piece of a specified length and orientation while the \syntax{custodial} mask checks for the presence of a ``custodial'' arrangement in which a line of pieces belonging to one player are ``sandwiched'' by a pair of pieces belonging to the other player. \methodname implements a wide range of low-level mechanics, and crucially these mechanics can be combined compositionally using first-order logic (excluding quantification) to form more complicated expressions. As an example, the condition
``\textit{if a player makes a line of 4 stones in a row or a 2 by 2 square of stones...}'' would be rendered as follows:
$$\texttt{(or (\textcolor{violet}{line} ``stone'' 4) (\textcolor{violet}{pattern} ``stone'' (\textcolor{violet}{square} 2)))}$$

\subsection{End Rules}

The last section of a game description in \methodname details the criteria that terminate a game. The \texttt{\textcolor{violet}{end}} section contains one or more ``end conditions'' which award a victory or loss to one to more players based on a binary predicate (e.g. forming a line or reaching a score threshold). A game can have multiple end conditions, which are then checked in order (so that win/loss conditions take precedence over draw checks, for instance).

\subsection{Design Considerations}
\label{sec:dsl_considerations}

While \methodname draws heavily from the \ludii description language, there are some important differences which go beyond just changes in syntax. Take \textit{Connect Four} as an example: \ludii aims to represent the game mechanically in a way that matches the natural language description of its rules as closely as possible.
Accordingly, the canonical representation of \textit{Connect Four} in \ludii features pieces that ``\texttt{Drop}'' into the ``\texttt{LastColumn}'' chosen by the player.
\methodname represents the action space differently -- players simply place a piece onto an empty board cell, with actions that are not directly above an existing piece or the bottom of the board marked as illegal. Mechanically, the two implementations of \textit{Connect Four} are identical -- the difference lies in how they are encoded (especially to simulated players or reinforcement learning agents).
While \methodname, like \ludii, strives to represent game descriptions intuitively, we primarily aim to provide a unified representation format across games so that general game-playing agents can more easily transfer knowledge and expertise from one game to another. As such, the size and shape of the action space for any game in \methodname is determined exclusively by the size of the board and the high-level play mechanics, regardless of how those mechanics might be translated into natural language.
This choice is also partially motivated by the specifics of working with the JAX library (see \Cref{sec:compiling}) and has implications for benchmarking and downstream use-cases (see \Cref{sec:benchmarking}).

\section{Compiling Game Descriptions into Game Environments}
\label{sec:compiling}

\methodname uses the \texttt{Lark} Python library to parse game descriptions into JAX environments, but the general approach is flexible enough to be used with different domains and parsing toolkits.
Broadly speaking, \methodname operates by defining the leaves of the grammatical parse tree (i.e. individual \texttt{masks}, \texttt{functions}, and \texttt{predicates}) as atomic functions written in JAX, which are then dynamically composed from the bottom-up to form higher-level operators used by the environment class. Consider again the following game expression:
$$\texttt{(or (\textcolor{violet}{line} ``stone'' 4) (\textcolor{violet}{pattern} ``stone'' (\textcolor{violet}{square} 2)))}$$


During compilation, the leaf-level nodes $\texttt{(\textcolor{violet}{line} ...)}$ and $\texttt{(\textcolor{violet}{pattern} ...)}$ are converted into JAX functions which map from the current game state to (in this case) a boolean truth value, and those functions are then passed up the parse tree. Higher-level nodes, such as \texttt{(and ...)}, receive the JAX functions corresponding to each of their children and return a \textit{new} JAX function that also takes the game state as input and implements the appropriate operation (in this case, boolean conjunction). In pseudocode, using the \texttt{Lark} library's \texttt{Transformer} paradigm, this looks like the following:
\begin{lstlisting}
def predicate_and(self, children):
    def predicate_fn(state):
        children_values = [child_fn(state) for child_fn in children]
        return all(children_values)

    return predicate_fn
\end{lstlisting}

In actuality, both the ``children functions'' and the combined ``predicate function'' must be written to be compatible with JAX's vectorization scheme and just-in-time (JIT) compilation. This imposes a number of implementational constraints, most notably that the size and shape of all arrays must be fixed at compile time. This means, for instance, that the dimensions of the ``legal action mask'' (and, hence, the size of the action space in general) cannot change as the game progresses and values like the number of iterations in a loop or the positions of a lookup mask must be pre-specified. Crucially, however, values that are determined during \textit{parsing} (such as the number of children for a given node, or the value of any arguments) can be safely passed into compiled JAX functions as static constants. This fact is what allows \methodname to dynamically create JAX functions that nonetheless obey the constraints of vectorization and JIT compilation. At the top of the parse tree, these composed JAX functions are ultimately used to define the behaviors that appear in the environment's \texttt{step} function, such as applying the player's action to the board and handling move effects.


\textbf{Precomputation: }
An important optimization for hardware accelerated environments is to express functions as batched matrix operations rather than iterative procedures. For instance, rather than checking for a line of pieces in \textit{Tic-Tac-Toe} by starting at the position of the last move and scanning out in each direction (as \ludii's implementation does), \pgx hard-codes the set of board indices that correspond to each possible line of three in the game (i.e. \texttt{[[0, 1, 2], [0, 3, 6], ...]}) and performs a single multi-dimensional index into the board array.
\methodname adopts and generalizes this approach: during parsing of \texttt{\textcolor{violet}{line}}, for example, the line indices are computed with respect to the size and shape of the game board (i.e. rectangular, hexagonal, ...) as well as the length and orientation of the desired line (i.e. diagonal, vertical, ...). Again, because these values depend only on attributes that are determined during parsing, they can be passed into JAX functions as constants and provide a substantial speedup when available.

\textbf{Dynamic State Attributes:} Different games require tracking different kinds of information about the current game state (such as scores or captured pieces). When \methodname compiles a game, it automatically extracts the attributes required to instantiate a game state and omits the others, thereby reducing the memory footprint of the entire state object. \methodname also automatically adds intermediary operations to each call of the environment's \texttt{step} function that help speed up later computations. For example, \textit{Hex} requires finding the connected components of the board to determine whether opposite edges are linked by a player's pieces. Rather than re-compute the connected components from scratch after each move, \methodname again adopts an approach from \pgx by detecting the presence of the \syntax{connected} function and maintaining a separate array of connected components that is locally updated after each move. This general procedure can be used to accommodate games with atypical or computationally expensive rules without affecting the runtime of other games.

\begin{figure}[t]
     \centering
     \includegraphics[width=\textwidth]{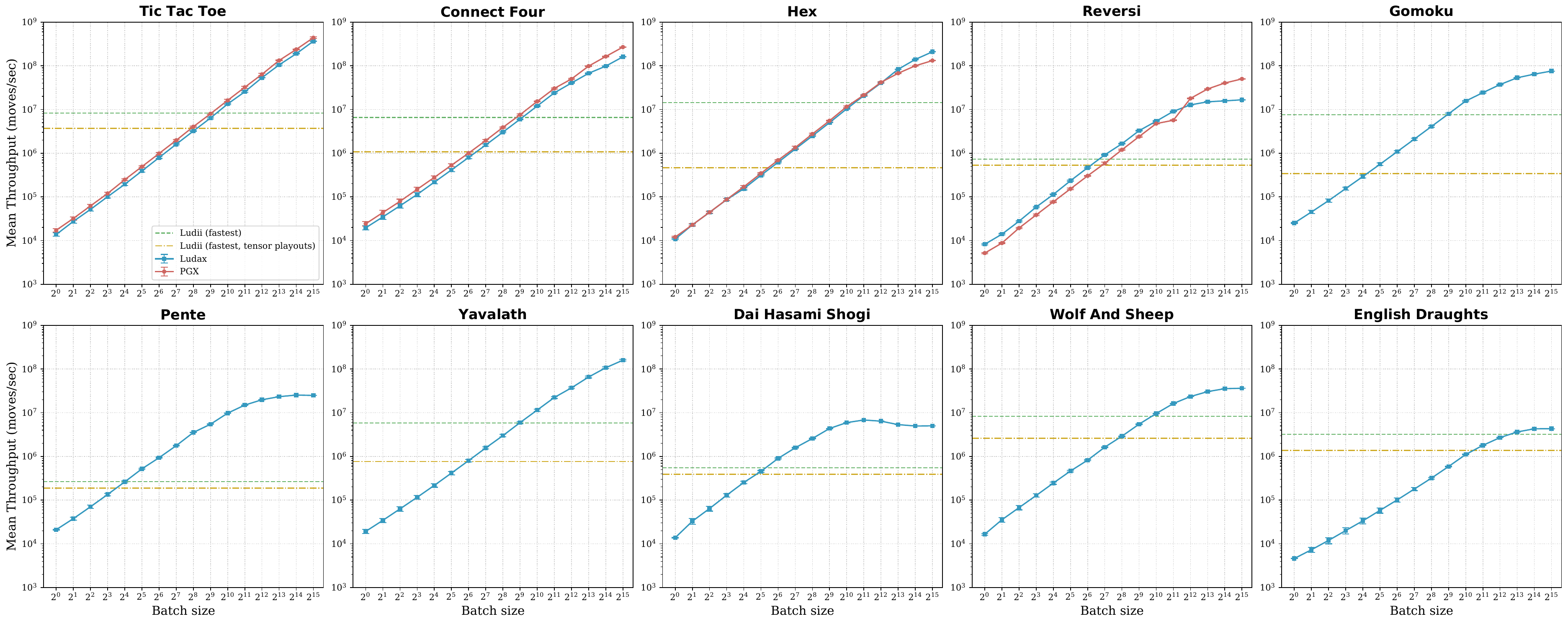}
     \caption{\textbf{Average throughput (moves per second) on various exemplar games for \methodname, \ludii, and \pgx.} The first four games are implemented in all three frameworks, while the remaining games are implemented only in \methodname and \ludii. Speeds for \methodname and \pgx are reported for 500 episodes of various batch sizes on a workstation with a single NVIDIA 4090 GPU and 36 threads on 18 CPU cores, while speeds for \ludii are the fasted recorded throughput for parallel execution on the same workstation across 1, 16, and 32 threads. Error bars are standard deviations over the 500 episodes.}
     \label{fig:throughput}
\end{figure}

\section{Expressive Range}
\label{sec:expressive_range}

\methodname supports a class of two-player, perfect-information board games played by placing, capturing, and moving game pieces of different types (and can be adapated to support simple gridworlds -- see \autoref{sec:gridworld}). Supported games range from the very simple (e.g. \textit{Tic-Tac-Toe}) to the relatively complex (e.g. \textit{English Draughts}, which features multiple kinds of pieces, forced captures, and sophisticated logic for granting extra turns). Existing games can also be easily modified, for instance by creating \textit{misère} variants or modifying the size and shape of the board. More importantly, the low-level mechanics used to implement individual games can be combined compositionally to form a vast array of unique interactions and dynamics. Further, because \methodname is a general description language and not a catalog of games, adding a single new mechanic or component expands the entire \textit{space} of representable games.

\section{Benchmarking}
\label{sec:benchmarking}

We benchmark the speed of \methodname on a set of 10 games, 4 of which are also implemented in both \ludii and \pgx (allowing for a full comparison) and 6 of which are implemented only in \methodname and \ludii. Again, we emphasize that these 10 games are just \textit{exemplars} of the class of games that \methodname supports, not an exhaustive list. A full description of each benchmark game is available in the Supplementary Material. We perform each of our benchmarking experiments on a workstation with a single NVIDIA 4090 GPU and an Intel i9-9980XE processor (36 threads on 18 CPU cores and 128GB of memory).

In \autoref{fig:throughput} we plot the throughput (in steps per second) under a uniformly random action policy for each game environment against the batch size (log scale on both axes), with the standard deviation of throughputs across episodes as error bars. \ludii supports parallelization via multi-threading: we report the fastest throughput on the same workstation when parallelized on 1, 16, and 32 threads. We note, however, that \ludii does not natively support reinforcement learning with fixed-size observation and action spaces (see \cite{Soemers_2022_DeepLearning} for a detailed discussion). To give a more realistic comparison in the context of RL specifically, we additionally report the throughput for \ludii when it constructs a fixed-size tensor for each state and indexes into a fixed-size approximation of the action space for each legal action.
Evaluations for \methodname and \pgx were obtained by performing 100 warmup full-game episodes at the specified batch size, followed by measuring the speed over 500 episodes, with each evaluation taking at most a few minutes to complete. Evaluations for \ludii were obtained by running warmup episodes for 10 seconds, followed by measuring the speed over 30 seconds of episodes.\footnote{We opted to measure speed for \methodname and \pgx using a fixed number of episodes because JAX's compilation procedure makes it difficult to halt execution after a specific elapsed wall time.} For games with potentially unbounded length (e.g. \textit{Dai Hasami Shogi}), we terminate games for both \methodname and \ludii after 200 total turns.

Overall, \methodname achieves speeds that are competitive with state-of-the-art JAX environments optimized for a single game. \pgx takes a slight edge in throughput at large batch sizes for \textit{Connect Four} and \textit{Reversi}, but \methodname remains within an order of magnitude (and even outspeeds \pgx in \textit{Hex}\footnote{Though \methodname's and \pgx's implementations differ slightly with the latter including the ``swap rule,'' this should not have a large effect on throughput under random playouts}).
\methodname also outspeeds \ludii across all 10 games, achieving a maximum speedup of between {\small $\sim$}3x (\textit{English Draughts}) and {\small $\sim$}450x (\textit{Hex}) in the tensor playout regime.
Some of this speedup is likely due to \methodname's comparatively smaller representation space, though 
\ludii also has optimized playout implementations tailored towards many of the categories of games covered by \methodname which might be difficult to apply in the context of deep learning \citep{Soemers_2022_Optimised}.

\section{Trained Agents}
\label{sec:rl}

\begin{figure}[t]
     \centering
     \includegraphics[width=0.49\textwidth]{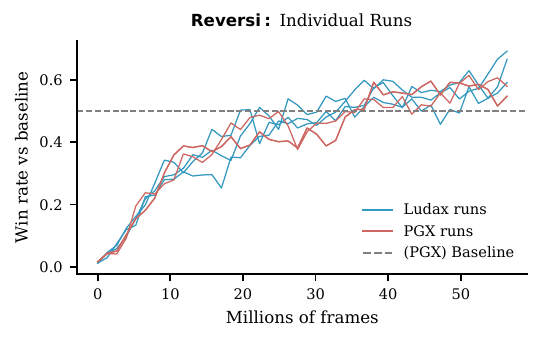}
     \includegraphics[width=0.49\textwidth]{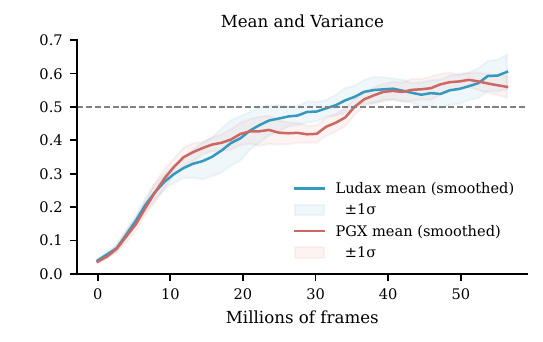}
     \caption{\textbf{Performance of reinforcement learning agents trained in the \methodname and \pgx implementations of \textit{Reversi} against the \pgx baseline agent.} On the left, we plot the average winrate of the learned agents against the baseline over time and across three separate runs. On the right, we plot the average and variance of the winrates. Each run took roughly 3 hours to complete on a workstation with a single A100 GPU.}
     \label{fig:reversi_reward}
\end{figure}

Finally, we demonstrate the feasibility of training reinforcement learning agents using the \methodname framework. We train our agent on the game \textit{Reversi} (also known as \textit{Othello}) using the \texttt{AlphaZero}-style \citep{silver2017mastering} training script from the \pgx library\footnote{https://github.com/sotetsuk/pgx/blob/main/examples/alphazero/train.py (used under Apache 2.0 license)}. We use the same ResNetV2 \citep{kaiming2016resnet}  network architecture and training hyperparameters as \pgx (full details available in the Supplementary Material) and train three separate runs on a single A100 GPU. Each run lasted roughly 57 million frames and took roughly three hours to complete.

We compare the performance of agents trained in the \methodname and \pgx environments against the baseline \textit{Reversi} agent provided by the \pgx library in \autoref{fig:reversi_reward}. Evaluations were performed by playing two batches of 1024 games (one with the learned agent as the first player and one as the second player), with actions sampled from the normalized output of the policy head at each step. We see that both learned agents achieve remarkably similar performances against the baseline, with little to no differences in learning speed or stability.
Although the \pgx implementation of the \textit{Reversi} environment is slightly more efficient, this translated into only marginal improvements in overall runtime (about $1.5\%$) owing to the shared overhead of network forward passes and weight updates. We also report training runs for Hex and Connect Four in \autoref{sec:morerlruns}. Like \pgx, \methodname offers a familiar API and an efficient set of implementations with which to train learned player agents.

\section{Limitations and Future Work}
\label{sec:limitations}

While we aim to continually expand the range of games expressible in \methodname, it will likely never match the full generality of a language like \ludii. As such, other frameworks may be more appropriate for use-cases in which a broad range of games is more important than rapid simulation. \methodname also does not support genres other than board games (e.g. video games, card games, ...) -- we leave the development of hardware accelerated description languages for such domains as an exciting area of future work. Compared to bespoke JAX implementations of board games (such as in the \pgx library), environments in \methodname have slightly worse throughput -- though the gap is marginal in a standard RL training setup. We deploy a number of optimizations to help close the efficiency gap when possible (see \Cref{sec:compiling}), but there are ultimately unavoidable trade-offs between speed and generality. For the purpose of training or benchmarking single-task agents on existing games, hard-coded simulators may remain the superior choice.

In light of these limitations, the most obvious avenues of extension for \methodname are the implementation and optimization of additional games and mechanics mechanics like irregular board shapes (e.g. \textit{Tabu Y}) and mechanically distinct gameplay phases (e.g. \textit{Nine-Men's Morris}). We also aim to provide a more robust visual interface for \methodname, both for the purpose of facilitating human-subject research \citep{carvalho2025nicewebrl} and the potential development of more ``human-like'' artificial agents which process the game board visually and select actions spatially. 

Looking further ahead, we are excited about the potential application of \methodname to the study of \textit{automatic game design}. Such systems depend on both a broad representation space and rapid evaluation of novel games -- see \autoref{sec:game-generation} for a preliminary investigation of \methodname's suitability for this kind of research. \methodname may also prove useful to research on \textit{human behavior and play}, for instance by expanding and accelerating work that models human player behavior \citep{zhang2024people}. Finally, \methodname offers an avenue to extend recent research in \textit{general game playing} (e.g. with large language models \citep{schultz2024mastering}) by providing a wide base of efficient game implementations that can in turn be leveraged for tree search algorithms or training world models.

\section{Conclusions}

We introduce a novel framework for games research that combines the generality of game description languages with the efficiency of modern hardware-accelerated learning environments. Our framework, \methodname, represents a broad class of two-player board games and compiles directly into code in the JAX Python library. Games in \methodname achieve speeds that are competitive with hand-crafted JAX implementations and faster than the widely-used \ludii game description language, and \methodname environments can easily be deployed in existing pipelines for deep reinforcement learning. Our framework helps widen and accelerate games research, with the potential to unlock new approaches in RL generalization, automatic game generation, and cognitive modeling.





\bibliography{bibliography, Dennis-Soemers-Bib}
\bibliographystyle{rlj}

\newpage
\appendix
\renewcommand{\thesubsection}{\Alph{subsection}}

\section{Example Games and Syntax}
\label{sec:dsl-examples}

\begin{figure}[h]
    \centering
    \includegraphics[width=\linewidth]{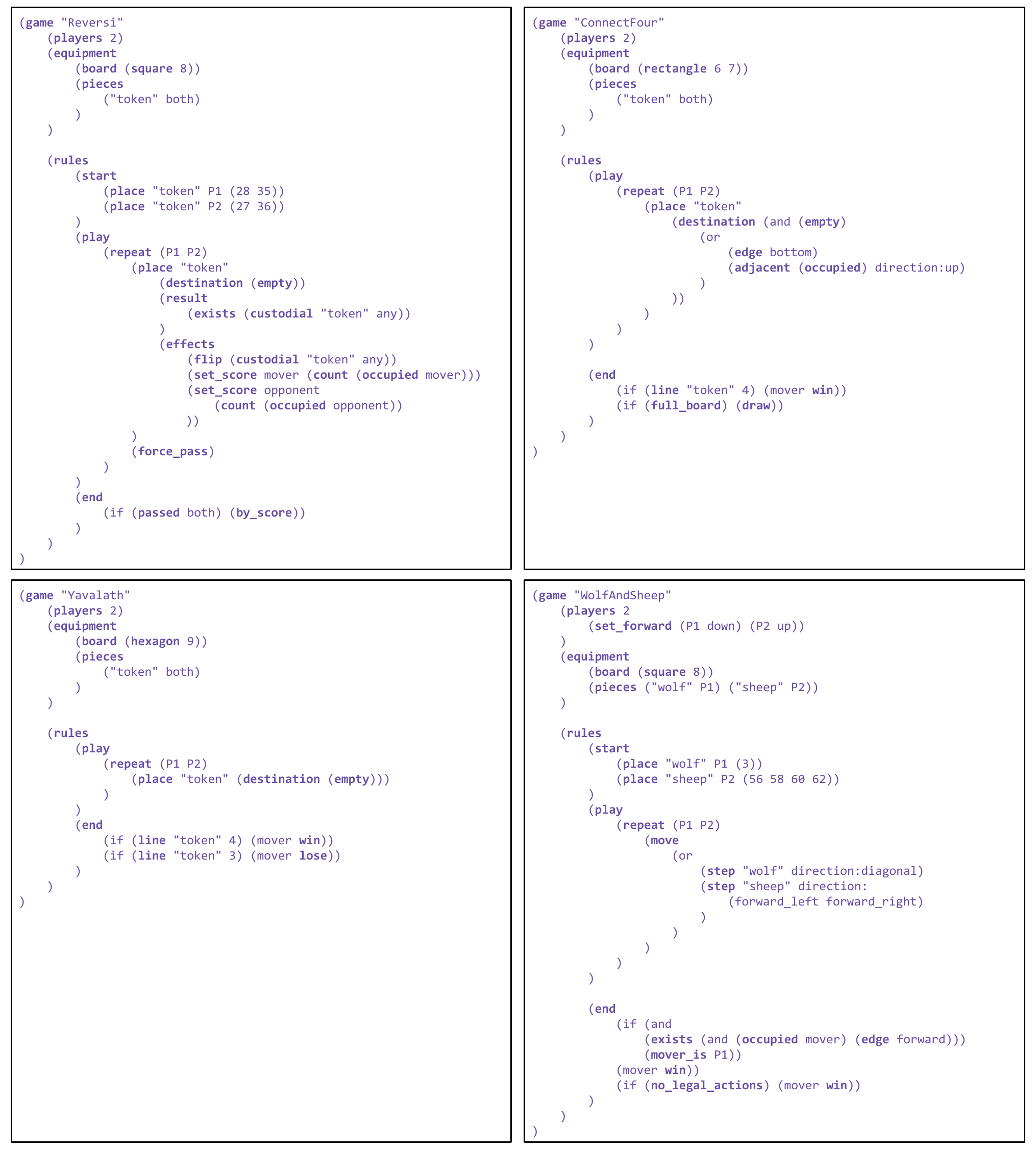}
    \caption{\textbf{\methodname syntax for \textit{Reversi}, \textit{Connect Four} (classic board games), \textit{Yavalath} and \textit{Wolf and Sheep}.} \methodname supports a wide range of games of vary complexity.}  
    \label{fig:dsl-example-app}
\end{figure}

In \autoref{fig:dsl-example-app} we present the \methodname syntax for a small set of exemplar games (\textit{Reversi}, \textit{Connect Four}, \textit{Yavalath}, and \textit{Wolf and Sheep}). Below, we present an annotated version of \textit{English Draughts} to help illustrate aspects of \methodname's syntax and structure.

\begin{lstlisting}[keywords={}, literate={place }{place }{1}{board}{board}{1}{end}{end}{1}{score}{score}{1}{empty}{empty}{1}{line}{line}{1}{opponent}{opponent}{1}{mover }{mover }{1}{any }{any }{1}{both}{both}{1}{square}{square}{1}{center}{center}{1}{hexagon}{hexagon}{1}{rectangle}{rectangle}{1}{win}{win}{1}{lose}{lose}{1}{P1 }{P1 }{1}{P2 }{P2 }{1}]

(game "English Draughts"
    
    // We define the "forward" direction for each player to simplify the movement logic
    (players 2
        (set_forward (P1up) (P2down))
    )

    // The game takes place on an 8-by-8 board and uses two kinds of pieces
    (equipment 
        (board (square 8))
        (pieces
            ("pawn" both)
            ("king" both)
        )
    ) 
    
    (rules

        // To begin, we place pawns for each player in a checkerboard pattern along the first and last two rows
        (start
            (place "pawn" P1 (40 42 44 46 49 51 53 55 56 58 60 62))
            (place "pawn" P2 (1 3 5 7 8 10 12 14 17 19 21 23))
        )
        
        (play
            // Players alternate making moves
            (repeat (P1 P2)

                // A move consists of one of the following options:
                (move
                    (or

                        // A pawn can hop over opposing pieces in either of the "forward" diagonals to capture them. This option has priority over non-capture moves
                        (hop "pawn" direction:(forward_left forward_right) hop_over:opponent capture:true priority:0)

                        // A pawn can step into unoccupied squares in the "forward" diagonals
                        (step "pawn" direction:(forward_left forward_right) priority:1)

                        // A king hops and captures in any diagonal direction...
                        (hop "king" direction:diagonal hop_over:opponent capture:true priority:0)

                        // ...and steps the same way
                        (step "king" direction:diagonal priority:1)
                    )
                    
                    (effects
                        // After moving, pawns on the forward edge of the board are promoted into kings
                        (promote "pawn" "king" (edge forward))

                        // If the player made a hop move (i.e. captured a piece) and the same piece could hop and capture again, the player gets an extra turn using that piece
                        (if (and (action_was mover hop) (can_move_again hop))
                            (extra_turn mover same_piece:true)
                        )
                    )
                )
            )
        )

        // If making a move results in the next player having no legal moves, then the most recent mover wins
        (end
            (if (no_legal_actions) (mover win))
        )
    )

    // For rendering, player one has the black pieces and player two has the white pieces
    (rendering
        (color P1 black)
        (color P2 white)
    )
)

\end{lstlisting}

\section{\methodname Grammar}
\label{sec:grammar}
Below we present the complete grammar specification for \methodname, using the syntax of the \texttt{Lark} Python library (raw string constants omitted for brevity).

\begin{lstlisting}[keywords={}, literate={place}{place}{1}{board}{board}{1}{end}{end}{1}{score}{score}{1}{empty}{empty}{1}{line}{line}{1}{opponent}{opponent}{1}{mover}{mover}{1}{any}{any}{1}{both}{both}{1}{square}{square}{1}{center}{center}{1}{hexagon}{hexagon}{1}{rectangle}{rectangle}{1}{win}{win}{1}{lose}{lose}{1}{P1}{P1}{1}{P2}{P2}{1}]

// ---Root---
game: "(" "game" name players equipment rules rendering? ")"

// ---Players---
players: "(" "players" positive_int forward_assignments? ")"
forward_assignments: "(" "set_forward" p1_assignment p2_assignment ")"
p1_assignment: "(" P1 (UP | DOWN | LEFT | RIGHT) ")"
p2_assignment: "(" P2 (UP | DOWN | LEFT | RIGHT) ")"

// ---Equipment---
equipment: "(" "equipment" board pieces regions? ")"
board: "(" "board" (board_square | board_rectangle | board_hexagon | board_hex_rectangle) ")"
?board_square: square_shape
?board_rectangle: rectangle_shape
?board_hexagon: hexagon_shape
?board_hex_rectangle: hex_rectangle_shape

// ---Pieces---
pieces: "(" "pieces" piece_definition+ ")"
piece_definition: "(" name (player_reference | BOTH) ")"
piece_reference: name

// ---Regions---
regions: "(" "regions" region_definition+ ")"
region_definition: "(" name (indices_arg | multi_mask_arg) ")"
region_reference: name

// ---Rules---
rules: "(" "rules" start_rules? play_rules end_rules ")"

// ---Start rules---
start_rules: "(" "start" start_rule+ ")"
start_rule: start_place
start_place: "(" "place" piece_reference player_reference (indices_arg | multi_mask_arg) ")"

// ---Play rules---
play_rules: "(" "play" play_phase+ ")"
play_phase: phase_once_through | phase_repeat
phase_once_through: "(" "once_through" play_mover_order play_super_mechanic ")"
phase_repeat: "(" "repeat" play_mover_order play_super_mechanic ")"
play_mover_order: "(" player_reference+ ")"

play_super_mechanic: play_mechanic force_pass?
play_mechanic: play_place | play_move
force_pass: "(" "force_pass" ")"

// ---Place rules---
play_place: "(" "place" piece_reference mover_reference? place_destination_constraint place_result_constraint? play_effects? ")"
place_destination_constraint: "(" "destination" super_mask ")"
place_result_constraint: "(" "result" super_predicate ")"

// ---Move rules---
play_move: "(" "move" move_type_definition play_effects? ")"
move_type_definition: move_type | "(" "or" move_type+ ")"
move_type: move_hop
         | move_slide
         | move_step

move_hop: "(" "hop" piece_reference direction_arg? piece_arg? hop_over_arg? capture_arg? priority_arg? ")"
move_slide: "(" "slide" piece_reference direction_arg? distance_arg? priority_arg? ")"
move_step: "(" "step" piece_reference direction_arg? priority_arg? ")"

// ---Effects---
play_effects: "(" "effects" play_super_effect+ ")"
play_super_effect: play_effect | play_conditional_effect
play_conditional_effect: play_if_effect | play_if_else_effect
play_if_effect: "(" "if" super_predicate play_effect ")"
play_if_else_effect: "(" "if" super_predicate play_effect "else" play_effect ")"

play_effect: effect_capture 
           | effect_extra_turn
           | effect_flip
           | effect_increment_score
           | effect_promote
           | effect_set_score

effect_capture: "(" "capture" super_mask mover_arg? increment_score_arg? ")"
effect_extra_turn: "(" "extra_turn" mover_reference same_piece_arg? ")"
effect_flip: "(" "flip" super_mask mover_arg? ")"
effect_increment_score: "(" "increment_score" mover_reference function ")"
effect_promote: "(" "promote" piece_reference piece_reference super_mask mover_arg? ")"
effect_set_score: "(" "set_score" mover_reference function ")"

// ---Functions---
function: function_add
        | function_connected
        | function_constant
        | function_count
        | function_line
        | function_multiply
        | function_pattern
        | function_score
        | function_subtract

function_add: "(" "add" function+ ")"
function_connected: "(" "connected" piece_reference multi_mask_arg mover_arg? direction_arg? ")"
function_constant: positive_int
function_count: "(" "count" super_mask ")"
function_line: "(" "line" piece_reference positive_int orientation_arg? exact_arg? player_arg? exclude_arg? ")"
function_multiply: "(" "multiply" function+ ")"
function_pattern: "(" "pattern" piece_reference (pattern_arg | shape_arg) rotate_arg? player_arg? exclude_arg? ")"
function_score: "(" "score" mover_reference ")"
function_subtract: "(" "subtract" function function ")"

// ---End rules---
end_rules: "(" "end" end_rule+ ")"
end_rule: "(" "if" super_predicate end_rule_result ")"
?end_rule_result: result_win | result_lose | result_draw | result_by_score

// -- Result definitions --
result_win: "(" (mover_reference | BOTH) "win" ")"
result_lose: "(" (mover_reference | BOTH) "lose" ")"
result_draw: "(" "draw" ")"
result_by_score: "(" "by_score" ")"

// -- Mask definitions --
super_mask: mask | mask_like_function | super_mask_and | super_mask_or | super_mask_not
super_mask_and: "(" "and" super_mask+ ")"
super_mask_or: "(" "or" super_mask+ ")"
super_mask_not: "(" "not" super_mask ")"

mask: mask_adjacent
    | mask_captured
    | mask_center
    | mask_column
    | mask_corners
    | mask_corner_custodial
    | mask_custodial
    | mask_edge
    | mask_empty
    | mask_hopped
    | mask_occupied
    | mask_prev_move
    | mask_promoted
    | mask_row
    | mask_region

mask_adjacent: "(" "adjacent" super_mask direction_arg? ")"
mask_captured: "(" "captured" ")"
mask_center: "(" "center" ")"
mask_column: "(" "column" nonnegative_int ")"
mask_corners: "(" "corners" ")"
mask_corner_custodial: "(" "corner_custodial" piece_reference mover_arg? ")"
mask_custodial: "(" "custodial" piece_reference custodial_length_arg mover_arg? orientation_arg? ")"
mask_edge: "(" "edge" (edge | FORWARD | BACKWARD) ")"
mask_empty: "(" "empty" ")"
mask_hopped: "(" "hopped" ")"
mask_occupied: "(" "occupied" mover_reference? ")"
mask_prev_move: "(" "prev_move" mover_reference ")"
mask_promoted: "(" "promoted" ")"
mask_row: "(" "row" nonnegative_int ")"
mask_region: "(" "region" region_reference ")"

// "Mask-like" functions are special functions that also support returning a mask. Currently this is only
// used for "line", but it might include other functions in the future
mask_like_function: function_line

// "Multi-masks" are special keywords that are manually split into multiple
// sub-masks at compile time. This is mostly useful for the "connected" function,
// which expects a list of masks to check for connections between
multi_mask: multi_mask_corners
          | multi_mask_edges
          | multi_mask_edges_no_corners

multi_mask_corners: "(" "corners" ")"
multi_mask_edges: "(" "edges" ")"
multi_mask_edges_no_corners: "(" "edgesNoCorners" ")"

// ---Predicate definitions---
super_predicate: predicate | super_predicate_and | super_predicate_or | super_predicate_not
super_predicate_and: "(" "and" super_predicate+ ")"
super_predicate_or: "(" "or" super_predicate+ ")"
super_predicate_not: "(" "not" super_predicate ")"

predicate: predicate_action_was
         | predicate_can_move_again
         | predicate_equals
         | predicate_exists
         | predicate_full_board
         | predicate_function
         | predicate_greater_equals
         | predicate_last_move_in
         | predicate_less_equals
         | predicate_mover_is
         | predicate_no_legal_actions
         | predicate_passed

predicate_action_was: "(" "action_was" mover_reference move_type_name ")"
predicate_can_move_again: "(" "can_move_again" move_type_name ")"
predicate_equals: "(" "=" function+ ")"
predicate_exists: "(" "exists" super_mask ")" // technically equivalent to (>= (count mask) 1) 
predicate_full_board: "(" "full_board" ")"
predicate_function: function // special syntax which is equivalent to "(>= function 1)"
predicate_greater_equals: "(" ">=" function function ")"
predicate_last_move_in: "(" "last_move_in" super_mask ")" // equivalent to (exists (and (prev_move mover) mask)), but more readable
predicate_less_equals: "(" "<=" function function ")"
predicate_mover_is: "(" "mover_is" player_reference ")"
predicate_no_legal_actions: "(" "no_legal_actions" ")"
predicate_passed: "(" "passed" (mover_reference | BOTH) ")"

// Additional (potentially optional) arguments for predicates
capture_arg: "capture:" boolean
custodial_length_arg: ANY | positive_int
direction_arg: "direction:" (direction | multi_direction)
distance_arg: "distance:" positive_int
exact_arg: "exact:" boolean
exclude_arg: "exclude:" multi_mask_arg
hop_over_arg: "hop_over:" (player_reference | mover_reference)
increment_score_arg: "increment_score:" boolean
mover_arg: "mover:" (mover_reference | BOTH)
multi_mask_arg: multi_mask | super_mask | "(" super_mask+ ")"
orientation_arg: "orientation:" orientation
pattern_arg: "(" positive_int indices_arg ")" // The positive int is the "width" of the pattern and the indices are the positions within the pattern
indices_arg: "(" nonnegative_int+ ")"
player_arg: "player:" (player_reference | mover_reference)
piece_arg: "piece:" piece_reference
priority_arg: "priority:" nonnegative_int
rotate_arg: "rotate:" boolean
same_piece_arg: "same_piece:" boolean
shape_arg: shape_definition

shape_definition: hexagon_shape
                | hex_rectangle_shape
                | rectangle_shape
                | square_shape

hexagon_shape: "(" "hexagon" odd_int ")"
hex_rectangle_shape: "(" "hex_rectangle" positive_int positive_int ")"
rectangle_shape: "(" "rectangle" positive_int positive_int ")"
square_shape: "(" "square" positive_int ")"

// Optional rendering details
rendering: "(" "rendering" rendering_detail+ ")"
rendering_detail: color_assignment
                | piece_shape_assignment

color_assignment: "(" "color" player_reference color ")"
piece_shape_assignment: "(" "shape" piece_reference piece_shape ")"

// General-purpose definitions
?number: SIGNED_NUMBER
?nonnegative_int: /(0|[1-9][0-9]*)/
?positive_int: /[1-9][0-9]*/
?odd_int: /[0-9]*[13579]/
?boolean: TRUE | FALSE
?edge: TOP | BOTTOM | LEFT | RIGHT | TOP_LEFT | TOP_RIGHT | BOTTOM_LEFT | BOTTOM_RIGHT
?direction: true_direction | relative_direction
?move_type_name: HOP | SLIDE | STEP
?multi_direction: "(" direction+ ")"
?true_direction: UP | DOWN | LEFT | RIGHT | UP_LEFT | UP_RIGHT | DOWN_LEFT | DOWN_RIGHT | VERTICAL | HORIZONTAL | ORTHOGONAL | DIAGONAL | BACK_DIAGONAL | FORWARD_DIAGONAL | ANY
?relative_direction: FORWARD | BACKWARD | FORWARD_LEFT | FORWARD_RIGHT | BACKWARD_LEFT | BACKWARD_RIGHT
?orientation: VERTICAL | HORIZONTAL | ORTHOGONAL | DIAGONAL | BACK_DIAGONAL | FORWARD_DIAGONAL | ANY
?color: WHITE | BLACK
?piece_shape: CIRCLE | SQUARE | TRIANGLE | STAR | DIAMOND
// ----------------------------

?player_reference: P1 | P2
?mover_reference: MOVER | OPPONENT
?name: STRING
variable_name: /\?[a-z][a-z0-9]*/
id: /[a-zA-Z0-9_]+/
\end{lstlisting}

\section{Benchmark Game Descriptions}
\label{sec:game-descriptions}
Below, we present natural language descriptions of the rules for each of the exemplar games analyzed in \Cref{sec:benchmarking}.

\textbf{\textit{Tic-Tac-Toe}:} Players take turns placing a piece into an empty space on a square 3-by-3 board. If a player forms a line of three of their pieces in a row (either vertically, horizontally, or diagonally), they win. If the board is completely full but no lines have been formed, then the game ends in a draw.

\textbf{\textit{Connect Four}:} Players take turns placing a piece into the top of one of the seven columns on a 6-by-7 board. The piece then ``falls'' until it rests on either the bottom of the board or another piece. A player can't place a piece into a column that is already ``full.'' If a player forms a line of four of their pieces in a row (either vertically, horizontally, or diagonally), they win. If the board is completely full but no lines have been formed, then the game ends in a draw.

\textbf{\textit{Hex}:} Players take turns placing a piece into an empty space on an 11-by-11 board composed of hexagonal tiles (forming a parallelogram, see visual depiction \href{https://en.wikipedia.org/wiki/Hex_(board_game)}{here}). The objective for the first player is to form a continuous path of their pieces that connects the top edge of the board with the bottom edge, while the objective for the second player is to do the same but connect the left and right edges of the board. The first player to achieve their objective wins the game. Because of the geometric properties of the board, it's not possible for the game to end in a draw.

\textbf{\textit{Reversi}:} The game takes place on a square 8-by-8 board. To begin, a white piece is placed at positions D4 and E5 and a black piece is placed at positions D5 and E4 (see visual depiction \href{https://en.wikipedia.org/wiki/Reversi#Rules}{here}). Players take turns placing a piece into an empty space such that a line of one or more of the opponent's pieces are ``sandwiched'' on either end by the player's pieces. This configuration is called a ``custodial'' arrangement of pieces. After placing a piece, any of the opponent's pieces which are in such a custodial arrangement are flipped and now belong to the player who just moved. It's possible for a single move to form multiple custodial arrangements in different directions, in which case all of the relevant pieces are flipped. If a player cannot make a legal move, they must pass (and they cannot pass without making a move otherwise). If both players pass, then the game is over. The winner is determined by the player who has the largest number of pieces on the board at the end of the game (in the event of a tie, the game ends in a draw).

\textbf{\textit{Gomoku}:} Players take turns placing a piece into an empty space on a square 15-by-15 board. If a player forms a line of exactly five of their pieces in a row (either vertically, horizontally, or diagonally), they win. However, forming a line of six or more does not count -- the player must have at least one line of exactly five. If the board is completely full but no lines of exactly five have been formed, then the game ends in a draw.

\textbf{\textit{Pente}:} Players take turns placing a piece into an empty space on a square 19-by-19 board. If a player forms a line of five of their pieces in a row (either vertically, horizontally, or diagonally), they win. In addition, if placing a piece causes a line of exactly two of the opponent's pieces to be put into a custodial arrangement, the two pieces are captured and removed from a board. Note that placing a piece \textit{into} a custodial arrangement formed by the opponent does not result in any pieces being captured. A player who captures at least 10 of the opponent's pieces over the course of the game wins. In the variant of \textit{Pente} implemented in \ludii and \methodname, the first player must make their first move into the exact center of the board.

\textbf{\textit{Yavalath}:} Players take turns placing a piece into an empty space on a regular hexagonal board with a diameter of 9 spaces. If a player forms a line of four of their pieces in any direction (either diagonally or horizontally\footnote{\methodname assumes a canonical orientation for hexagonal boards in which the diameter stretches from left to right, though it is functionally equivalent to the orientation in which the diameter runs vertically)}), they win. However, if a player forms a line of three of their pieces in a row without also forming a line of four, they lose. If the board is completely full but no lines of four or three have been formed, then the game ends in a draw.


\textbf{\textit{Dai Hasami Shogi}:} The game takes place on a square 9-by-9 board. To begin, white pieces are placed on the bottom two rows of the board and black pieces are placed on the top two rows. Players take turns moving one of their pieces, either by sliding it any number of squares vertically or horizontally (i.e. as a rook) or by hopping over one piece (belonging to either player) vertically or horizontally into an empty square. Hopping over a piece does not capture it, but opposing pieces can be captured ``custodially'' (i.e. by moving to surround an enemy piece on both sides vertically or horizontally). An opponent's piece in a corner can also be captured by moving a piece to occupy both orthogonally-adjacent squares. A player wins if they manage to form a horizontal or vertical line of 5 pieces in a row if none of those pieces are in their starting rows.


\textbf{\textit{Wolf and Sheep}:} The game takes place on a square 8-by-8 board. Player 1 begins with four ``sheep'' pieces placed on the dark squares of the bottom row while Player 2 begins with a single ``wolf'' piece placed on one of the center dark squares in the top row. Players take turns moving one of their pieces to an unoccupied square -- ``sheep'' can move in either of the forward diagonal directions (i.e. towards the top of the board) while the ``wolf'' can move in any diagonal direction. Player 1 wins if they manage to block all of Player 2's legal actions while Player 2 wins if they manage to get their piece to the bottom edge of the board.

\textbf{\textit{English Draughts}:} Also known as \textit{Checkers}, this game takes place on a square 8-by-8 board. Player 1 and 2 with 12 ``pawns'' placed on the dark squares of the bottom three and top three rows, respectively. A pawn can move by either stepping into an unoccupied square in one of the forward diagonals (i.e. towards the top of the board for Player 1 and towards the bottom of the board for Player 2) or by hopping over an opponent's piece in one of those directions and capturing it. If a player has the option to make a capture move, they must do so. In addition, if after making a capture, the capturing piece could make another capture then the active player gets another turn in which they must move their previously-moved piece. If a ``pawn'' reaches the opposite edge of the board, it is promoted to a ``king'' which can move and capture in any diagonal direction. The game ends when one player has no legal actions, at which point they lose.

\section{Gridworld Representation}
\label{sec:gridworld}

\begin{figure}[t]
    \centering
    \includegraphics[width=0.5\linewidth]{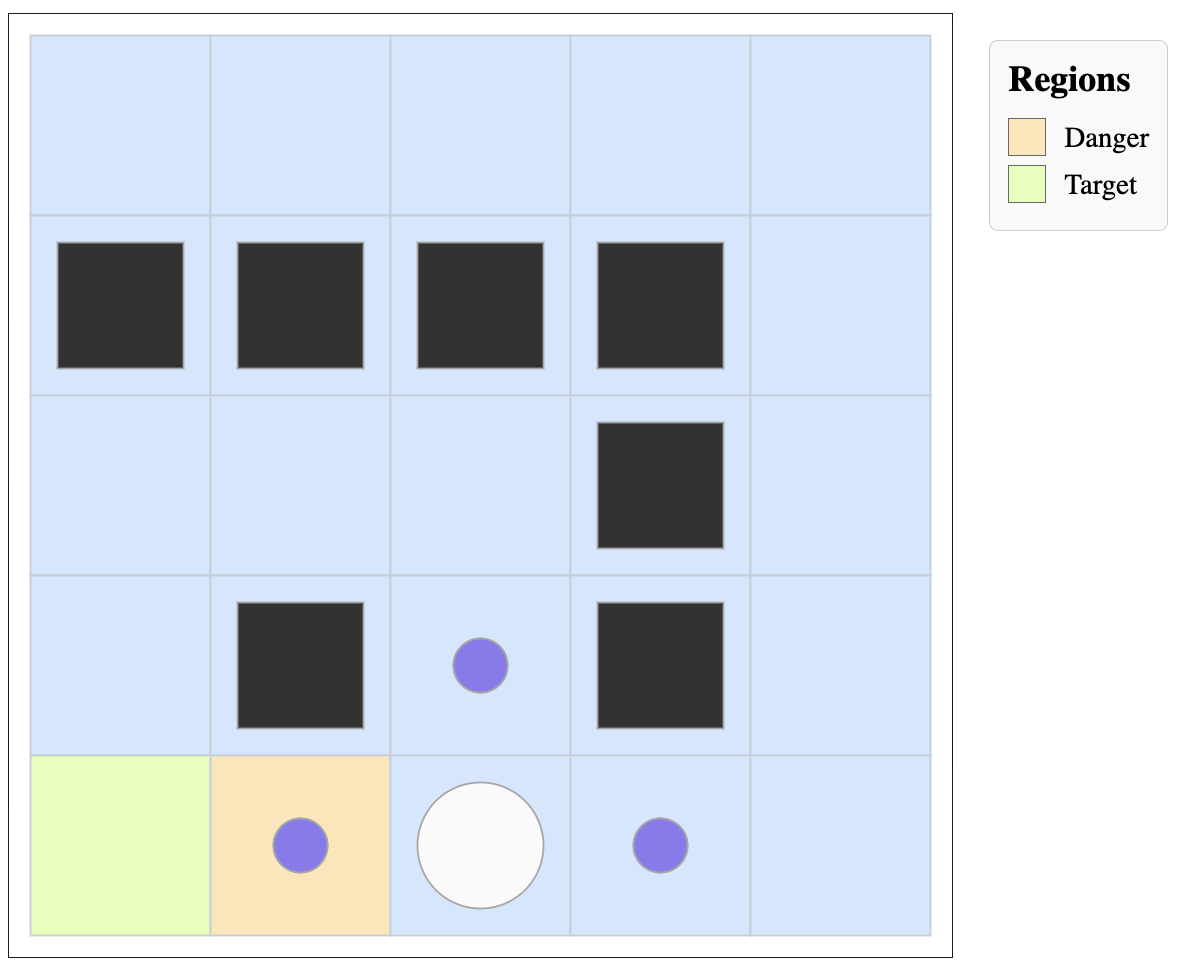}
    \caption{\textbf{\methodname rendering of a simple gridworld environment akin to \texttt{FrozenLake}.} \methodname's syntax can be adapted to represent single-player games with movement dynamics more typical of simple video games. Player 1 (white) moves a single piece (circle) one step at a time, attempting to avoid the ``danger'' region in orange and reach the ``target'' region in green.}
    \label{fig:gridworld-example}
\end{figure}

While \methodname is designed to primarily support two-player board games, its flexible syntax allows it to encode certain kinds of simple single-player games often used as RL training environments. By giving one player a single piece capable of stepping one square orthogonally into empty board positions, defining certain board regions as either ``targets'' or ``dangers,'' and omitting the second player from the move order definition, the resulting game effectively encodes a simple gridworld like the classic \texttt{FrozenLake} environment. While the resulting action space is not exactly analogous to a traditional gridworld (i.e. for the \methodname game, the action space would be defined based on the number of legal squares), it's very straightforward to detect the specification of this kind of game at parse time and adjust the underlying observation and action space accordingly. An example of such an environment rendered using \methodname's interactive debugger is provided in \autoref{fig:gridworld-example}.

\section{Training Hyperparameters}
\label{sec:hyperparams}

Below we provide the exact training hyperparameters used in the reinforcement learning experiments in \Cref{sec:rl}. These are largely copied from the \pgx implementation. 

\begin{itemize}
    \item \textbf{Model architecture:} \texttt{ResnetV2}
    \item \textbf{Number of channels:} 128
    \item \textbf{Number of layers:} 6
    \item \textbf{Self-play batch size:} 1024
    \item \textbf{Self-play simulations:} 32
    \item \textbf{Self-play max steps:} 256
    \item \textbf{Training batch size:} 4096
    \item \textbf{Learning rate:} 0.001
    \item \textbf{Evaluation frequency:} 5
    \item \textbf{Training iterations:} 219
\end{itemize}

Note that each ``iteration'' consists of generating play data for 256 steps using the self-play batch size of 1024 (see \cite{koyamada2023pgx}). We train the model for 219 iterations, which corresponds to $256 \times 1024 \times 219 = 57409536$ (or roughly $57$ million) steps in the environment.

\section{Additional Training Runs}
\label{sec:morerlruns}
\begin{figure}[t]
     \centering
     \includegraphics[width=0.49\textwidth]{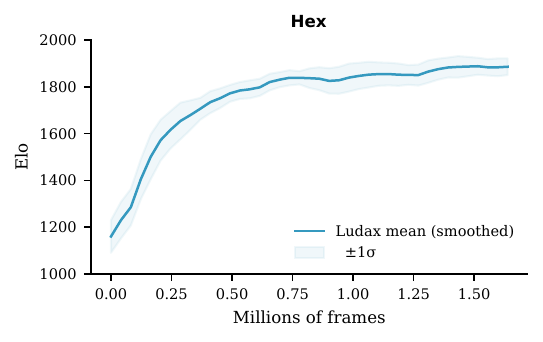}
     \includegraphics[width=0.49\textwidth]{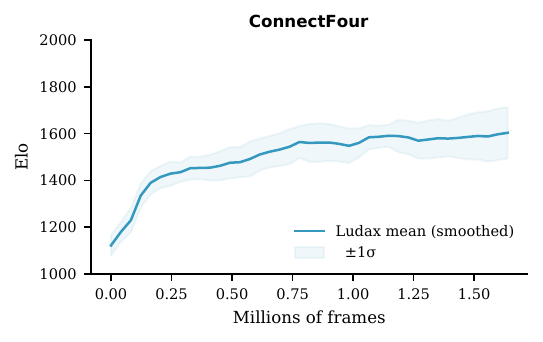}
     \caption{\textbf{Performance of reinforcement learning agents trained in the \methodname.} Elo scores are computed between 40 model checkpoints and averaged over three runs. Each run took roughly 1 hour to complete on four GH200s.}
     \label{fig:elo}
\end{figure}

In addition to the head-to-head comparison between \methodname and \pgx, we present the results of training additional agents in \methodname's Hex and Connect Four environments in \autoref{fig:elo}. Owing to differing state and action space representations, we cannot directly compare \methodname agents to \pgx's baseline by having the agents play against each other. Instead, we report the base-10 Elo score (similarly to \cite{David2018}) of 40 model checkpoints across training, averaged over three runs. We perform 64 playthroughs for each pairing of checkpoints until all Elo ratings converge. While the specific Elo scores are not meaningful in isolation, the curves demonstrate that standard RL techniques can learn effectively in a variety of \methodname environments.

\section{Game Generation}
\label{sec:game-generation}

We attempt to synthesize new games in the \methodname DSL using two approaches: random sampling and LLM-based generation. In Table \ref{tbl:gavel-scores}, we present the \texttt{GAVEL} game evaluation metrics for each method.

\textbf{Random Sampling:} Games are generated by naive uniform random sampling. Starting from the root game ``ludeme'' (i.e. production rule), we sample the next ludeme among those which are valid continuations according to the grammar. Additionally, we impose a maximum syntax tree depth of 5, beyond which a closing bracket is always given priority.

\begin{lstlisting}[
frame=single,
float,
caption={System instruction for LLM-based generation.},
captionpos=b,
label={lst:system-instruction},
keywords={},
literate={place}{place}{1}{board}{board}{1}{end}{end}{1}{score}{score}{1}{empty}{empty}{1}{line}{line}{1}{opponent}{opponent}{1}{mover}{mover}{1}{any}{any}{1}{both}{both}{1}{square}{square}{1}{center}{center}{1}{hexagon}{hexagon}{1}{rectangle}{rectangle}{1}{win}{win}{1}{lose}{lose}{1}{P1}{P1}{1}{P2}{P2}{1}]
Invent simple rules for a novel two player abstract strategy game
called {name}. Implement it in the ludax language. You will find
attached the ludax's grammar as well as a few examples of games
implemented in ludax. Start by implementing a simplified version
of your rules, and then incrementally add rules that are harder to
express in ludax. At each step, make sure you write a compilable
game according to ludax's grammar.
\end{lstlisting}

\begin{table}[t]
\centering
\caption{GAVEL-based evaluation metrics for 100 generated games, obtained either by uniform random sampling or an LLM. As a baseline, we report results for all default games in Appendix \ref{sec:game-descriptions}. \textit{Playable} and \textit{Interesting} denote percentages over all generated games (\textit{Playable} $\geq$ \textit{Interesting}). \textit{GAVEL score} and \textit{Strategic Depth} report the median and standard deviation, computed only on playable games.}
\begin{tabular}{lrrrr}
\toprule
\textbf{Method} & \textbf{Playable} & \textbf{Interesting} & \textbf{GAVEL Score} & \textbf{Strategic Depth}\\
\midrule
Default Games & 100\% & 100\% & 0.69 {\small $\pm 0.15$} & 0.66 {\small $\pm 0.15$} \\
Random Sampling & 4\% & 0\% & 0.00 {\small $\pm 0.00$} & 0.00 {\small $\pm 0.00$} \\
GPT-OSS-120B & 95\% & 83\% & 0.59 {\small $\pm 0.22$} & 0.58 {\small $\pm 0.17$} \\
LLaMa-4-17B & 82\% & 42\% & 0.49 {\small $\pm 0.21$} & 0.68 {\small $\pm 0.23$} \\
\bottomrule
\end{tabular}
\label{tbl:gavel-scores}
\end{table}

\textbf{LLM-based Generation:} Games are generated as a few-shot task. The model is prompted with a system instruction (Listing \ref{lst:system-instruction}), the full grammar (Appendix \ref{sec:grammar}), and the game implementations from Appendix \ref{sec:game-descriptions} as examples. The model is instructed to describe the rules of a new game and produce multiple \methodname implementations of increasing complexity; we evaluate only the final game produced. To encourage diversity, each attempt is seeded with a randomly generated and nonsensical game name such ``Outstanding Rainbow Spaniel.''

\textbf{GAVEL-like Evaluation:} Inspired by \citet{todd2024gavel}, we assess each generated game as follows:
\begin{enumerate}
    \item A game is \textit{playable} if its description compiles and runs without error.
    \item For each \textit{playable} game, we run agent-vs-agent playthroughs using a custom JAX implementation of MCTS with UCB1 \citep{Kocsis_2006_Bandit}.
    \item We compute the following heuristics from these playthroughs: 
    \begin{itemize}
      \item \textbf{Balance:} max winrate gap between players  
      \item \textbf{Decisiveness:} fraction of non-draw outcomes  
      \item \textbf{Completion:} fraction of games reaching a terminal state  
      \item \textbf{Agency:} fraction of turns with $>1$ legal move  
      \item \textbf{Coverage:} fraction of board sites occupied at least once
      \item \textbf{Strategic Depth:} difference in winrate between a stronger MCTS agent and a weaker one (fewer simulations).
    \end{itemize}
\end{enumerate}

The overall ``\texttt{GAVEL} score'' is the harmonic mean of the individual heuristic scores. Games with a \texttt{GAVEL} score $>0.4$ are deemed potentially \textit{interesting}. We note that this experiment is preliminary: it omits diversity measures, and the limited search budget for MCTS means they will frequently miss good moves a stronger agent might find. Nevertheless, the fact that an LLM can implement novel games in \methodname without finetuning suggests that \methodname’s grammar is intuitive and highlights its potential for game generation.

\textbf{Hyperparameters:} For each method, we sample 100 games. For the LLM-based methods, we use a sampling temperature of 0.2. To compute the evaluation score, we run 100 agent-vs-agent simulations for each game. The MCTS agents perform 100 iterations (i.e. traversal, expansion, and random rollout) for each action. For the ``strategic depth'' evaluation we compare against an MCTS agent that performs 50 iterations per action.



\end{document}